\documentclass[letterpaper, 10 pt, conference]{ieeeconf}  % Comment this line out if you need a4paper

\IEEEoverridecommandlockouts                              % This command is only needed if 
\usepackage{times}
\usepackage{soul}
\usepackage{url}
\usepackage[hidelinks]{hyperref}
\usepackage[utf8]{inputenc}
\usepackage[small]{caption}
\usepackage{graphicx}
\usepackage{amsmath}
\usepackage{booktabs}
\usepackage{algorithm}
\usepackage{algorithmic}
\usepackage[switch]{lineno}
\usepackage{multirow}
\usepackage{ulem}

\usepackage{amssymb,amsfonts}
\usepackage{textcomp}
\usepackage{xcolor}
\usepackage{multirow}
\usepackage{array}
\usepackage[dvipsnames,table,xcdraw]{xcolor}
\usepackage{comment}
\usepackage{makecell}
\usepackage{pdfpages}
\usepackage{adjustbox}

\title{\LARGE \bf
Aligning LLM+PDDL Symbolic Plans with Human Objective Specifications through Evolutionary Algorithm Guidance*
}

\author{Owen Burns$^1$, Dana Hughes$^2$, and
Katia Sycara$^2$% <-this % stops a space
\thanks{*This project was funded by ONR Award No. N000142312840 and NSF Award No. 2021-67021-35329.}% <-this % stops a space
\thanks{
$^{1}$College of Engineering and Computer Science, University of Central Florida, Orlando, FL, USA
        {\tt\small ow446044@ucf.edu}}%
\thanks{$^{2}$Robotics Institute, Carnegie Mellon University, Pittsburgh, PA, USA
        {\tt\small danahugh@andrew.cmu.edu,katia@cs.cmu.edu}}%
}

\begin{document}

\maketitle
\thispagestyle{empty}
\pagestyle{empty}

\begin{abstract}

Automated planning using a symbolic planning language, such as PDDL, is a general approach to producing optimal plans to achieve a stated goal.  However, creating suitable machine understandable descriptions of the planning domain, problem, and goal requires expertise in the planning language, limiting the utility of these tools for non-expert humans. Recent efforts have explored utilizing a symbolic planner in conjunction with a large language model to generate plans from natural language descriptions given by a non-expert human (LLM+PDDL). Our approach performs initial translation of goal specifications to a set of PDDL goal constraints using an LLM; such translations often result in imprecise symbolic specifications, which are difficult to validate directly.  We account for this using an evolutionary approach to generate a population of symbolic goal specifications with slight differences from the initial translation, and utilize a trained LSTM-based validation model to assess whether each induced plan in the population adheres to the natural language specifications.  We evaluate our approach on a collection of prototypical specifications in a notional naval disaster recovery task, and demonstrate that our evolutionary approach improve adherence of generated plans to natural language specifications when compared to plans generated using only LLM translations. The code for our method can be found at https://github.com/owenonline/PlanCritic.

\end{abstract}

%---------------------------------------------------------------------------------

\section{INTRODUCTION}\label{sec:introduction}
%The ability of human planners to effectively manage situations involving complex, multifaceted objectives is limited. We consider the case of a non-expert human planner in particular. In an increasingly interconnected world the potential downstream impacts of failed planning only grow, a fact borne out by the worldwide supply chain shortages seen in the wake of COVID \cite{duong_ripple_2023}.

Automated symbolic planners are well-established, domain-independent tools for generating optimal plans from domain, problem, and goal descriptions defined in in a formal language, 
%Formally defining tasks, domains, and goals in a symbolic language, 
such as the Planning Domain Definition Language (PDDL).%, allows for automating plan generation from provided descriptions. 
While such planners %carrying out that generation
excel at solving complex tasks, %the strict format in which problems must be specified
the need for expertise in the formal language typically makes these systems impractical for use by non-experts, such as decision makers or mission commanders~\cite{boddy_imperfect_2003}.  While large language models (LLMs) have demonstrated proficiency at translating natural language to PDDL expressions, such as goal descriptions~\cite{xie_translating_2023}, the resulting specifications often suffer from syntactical mistakes or deviate from the user's intent~\cite{gragera_exploring_nodate}. 

Recent research into utilizing LLMs for generating task descriptions in PDDL have focused primarily on the ability of an LLM to generate accurate PDDL from natural language; in this paper, we focus instead on developing an LLM-based system for collaborative human-AI planning.  In such a setting, plan generation may involve multiple iterations of human feedback to an AI in order to effectively capture the human's preferences for various plan constraints.  Approaches focused purely on LLM-based PDDL generation also lack the ability to search the space of available plans beyond what may be arrived at downstream of the LLM's generation; we address that limitation by using a genetic algorithm to efficiently explore the planning space. 

In this paper, we introduce a neurosymbolic framework capable of assisting human planners in dynamic and complex environments by generating state trajectory constraints which generate updated plans, based provided user feedback in natural language to an existing plan.  Our system uses a genetic algorithm-based optimizer to search the space of symbolic plan specifications, based on an initial specification provided by a possibly erroneous LLM translation.  A specification adherence model to estimate the degree to which plans adhere to natural language user feedback.  Finally, we demonstrate that our system can improve the alignment of plans generated from user feedback, compared to plans produced using only LLM translations of user feedback.

The remainder of the paper is organized as follows.  Section~\ref{sec:related_work} describes recent work on using LLMs in symbolic planning scenarios.  Section~\ref{sec:background} provides a brief overview of PDDL and Genetic Algorithms.  Section~\ref{sec:technical_approach} describes the overall technical approach, and provides details of each component of the framework.  Section~\ref{sec:evaluation} provides a description of the naval disaster response scenario used to evaluate our approach, the results of the evaluation are detailed in Section~\ref{sec:results}.  We provide a conclusion and highlight potential future research efforts in Section~\ref{sec:conclusion}.

\section{RELATED WORK}\label{sec:related_work}

Interest in combining the flexibility of neural models with the speed and correctness of classical planners has been extensively studied. Early works focused on learning domains, with~\cite{miglani_nltopddl_nodate} using a deep-q network to translate existing instruction manuals into domains one-shot and~\cite{lamanna_online_2021} going a step further to learn STRIPS domains online with the help of an agent generating informative plan traces for training. 

With the advent of LLMs and their effectiveness at translation, particularly from natural language into (PDDL)~\cite{xie_translating_2023}, focus shifted towards verification, with~\cite{smirnov_generating_2024} using an LLM-based agent to test the self-consistency of generated domains and make necessary repairs. The linguistic proficiency brought by LLMs allowed research into neurosymbolic planning architectures to branch out from domain learning as well. One approach explored is to generate a symbolic description of some aspect of the planning problem from natural language, such as~\cite{dagan_dynamic_2023} or~\cite{liu_llmp_2023}, which would then be provided to an existing symbolic planner to produce a symbolic plan. Alternatively, the LLM could be queried for a symbolic plan directly from a natural language description of the problem, and which could then be symbolicly verified in an post-hoc manner~\cite{capitanelli_framework_2024}. Other works took advantage of the coding proficiency of larger models, with \cite{silver_generalized_2023} using GPT-4 few-shot to generate programs which in turn created valid plans from a domain. These approaches, however, are all limited in their ability to explore the space of available plans due to their reliance on the LLM for variation.

% In the context of planning and decision-making, reinforcement learning from human feedback (RLHF) has been employed to refine the outputs of planning algorithms based on user feedback. For example, \cite{christiano2023deepreinforcementlearninghuman} demonstrated the use of RLHF in Atari games, where human feedback was used to train an agent to perform better than with traditional reward functions alone. Similarly, in robotic manipulation tasks, RLHF has been used to teach robots complex behaviors that are difficult to specify through conventional reward functions \cite{hiranaka2023primitiveskillbasedrobotlearning}.

\section{BACKGROUND}\label{sec:background}

\subsection{Symbolic Planning}
Symbolic planning is a domain-independent automated approach to generating a sequence of steps in order to perform a task, given a symbolic description of the state of the world, a set of actions with corresponding preconditions and postconditions, and a goal description.  Multiple tools exist to automate planning given a symbolic task description.

Planning Domain Definition Language (PDDL)~\cite{aeronautiques1998pddl} is a common domain-independent language for symbolically specifying planning domains and problems.  PDDL is used to express \textit{domain descriptions} and \textit{problem descriptions}.  A domain description is used to define object types, predicates used for logical expression of facts about world states, and actions; actions are parameterized by variables indicating objects involved with the action, and a set of preconditions (predicates which must be true to make the action valid) and postconditions (predicates that are true or false after the action).  The problem description is used to define the initial state of the world in terms of a set of predicates, object instances, and a goal description as a logical predicate which must be true.  PDDL3~\cite{gerevini2006preferences} introduced \textit{state-trajectory constraint} predicates to express temporal constraints about a plan's trajectory, e.g., indicating that a predicate must always be true, or must be true sometime during the generated plan.

\subsection{Genetic Algorithms}
Genetic algorithms (GAs) are a family of metaheuristic algorithms inspired by the principles of natural selection and genetics \cite{goldberg_genetic_1989}, which have been utilized in several domains, including production scheduling \cite{neumann_genetic_2024}, route optimization \cite{baker_genetic_2003}, and resource allocation \cite{alcaraz_robust_2001}.  GAs are well suited to optimizing ``black-box'' problems, where derivatives of the objective function of the problem may not exist, or may be prohibitively expensive to calculate~\cite{mitchell_introduction_1996}.

Genetic algorithms maintain and evolve a population of candidate solutions; each individual in the population consists of a \textit{genotype}, which encodes a unique solution of the problem, referred to as the individual's \textit{phenotype}.  The individual's \textit{fitness} is determined by an objective function applied to its phenotype.  Genetic algorithms iteratively improve the population of candidate solutions through multiple generations.  Individuals in a new generation are generated by stochastically selecting two \textit{parents} from the current population, based on their fitness; the \textit{child} is generated by performing a \textit{crossover} operation on the parents' genotypes, followed by a \textit{mutation} operation on the resulting genotype. 

\section{TECHNICAL APPROACH}\label{sec:technical_approach}
\begin{figure}[!h]
  \centering  \includegraphics[width=1\linewidth]{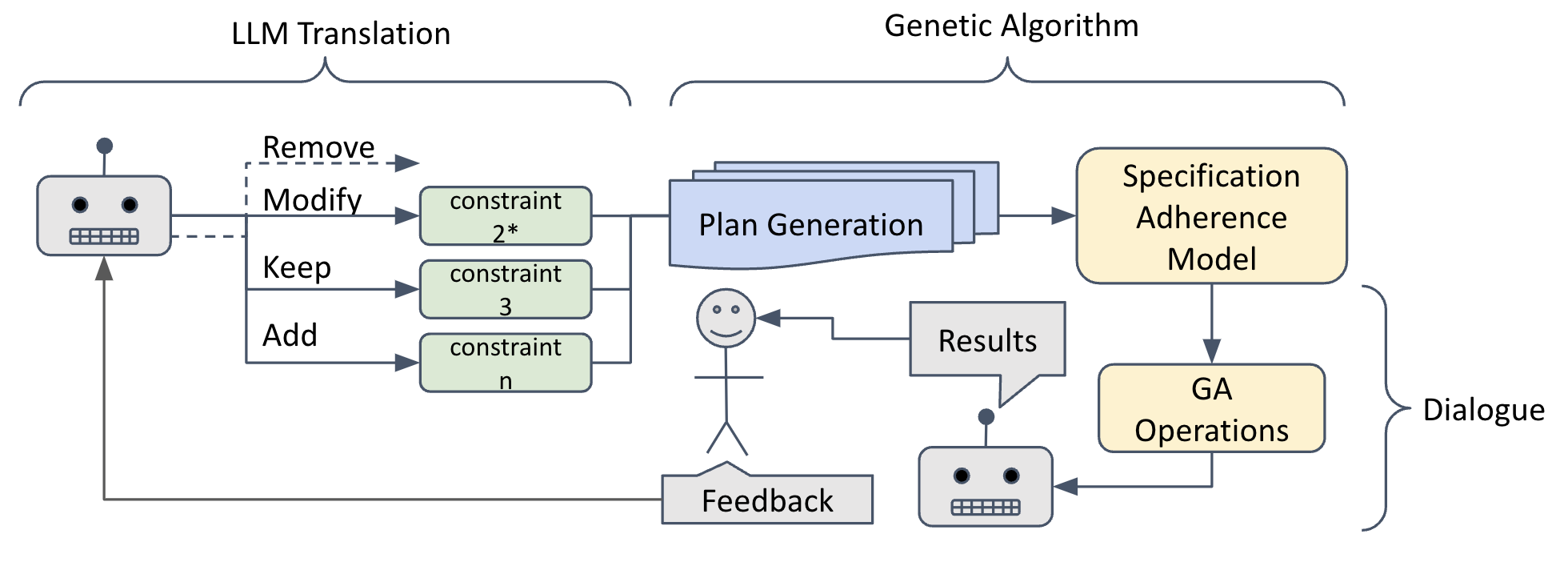}
  \vspace{-.5cm}
  \caption{Diagram of the proposed system. Natural language feedback to an initial plan provided by the user are translated to an initial plan specification.  The genetic algorithm searches locally over the initial plan specification, producing a population of candidate specifications.  The specification adherence model quantifies the extent to which plans adhere to the user's natural language feedback, and is used to guide the genetic algorithm search.}
  \label{fig:system_diagram}
\end{figure}

The aim of our system is to collaborate with a user to generate a symbolic plan whose specifications adhere to the preferences of the user, through feedback from the user in natural language.  Our system, shown in Figure~\ref{fig:system_diagram}, consists of the following components:  a \textit{user interface and large language model (LLM)} (detailed in Section~\ref{sec:user_interface}), a \textit{Symbolic Planner} (detailed in Section~\ref{sec:planner}), a \textit{Specification Adherence Model} (detailed in Section~\ref{sec:adherence_model}), and a \textit{Genetic Algorithm Component} (detailed in Section~\ref{sec:genetic_algorithm}). 

The system uses an LLM to translate a set of symbolically grounded natural language feedback statements from the user to a symbolic plan specification.  The symbolic planner generates a plan based on a plan specification.  The LLM is prone to mistranslating one or more statements in the user feedback, resulting in plans which fail to fully adhere to the user's feedback.  The system quantifies this adherence using the Specification Adherence Model.  In order to improve the adherence rate of generated models, the system uses the Genetic Algorithm Component to search over symbolic plan specifications to correct mistranslations.

\subsection{User Interaction}\label{sec:user_interface}
\begin{figure}[!h]
  \centering  \includegraphics[width=0.9\linewidth]{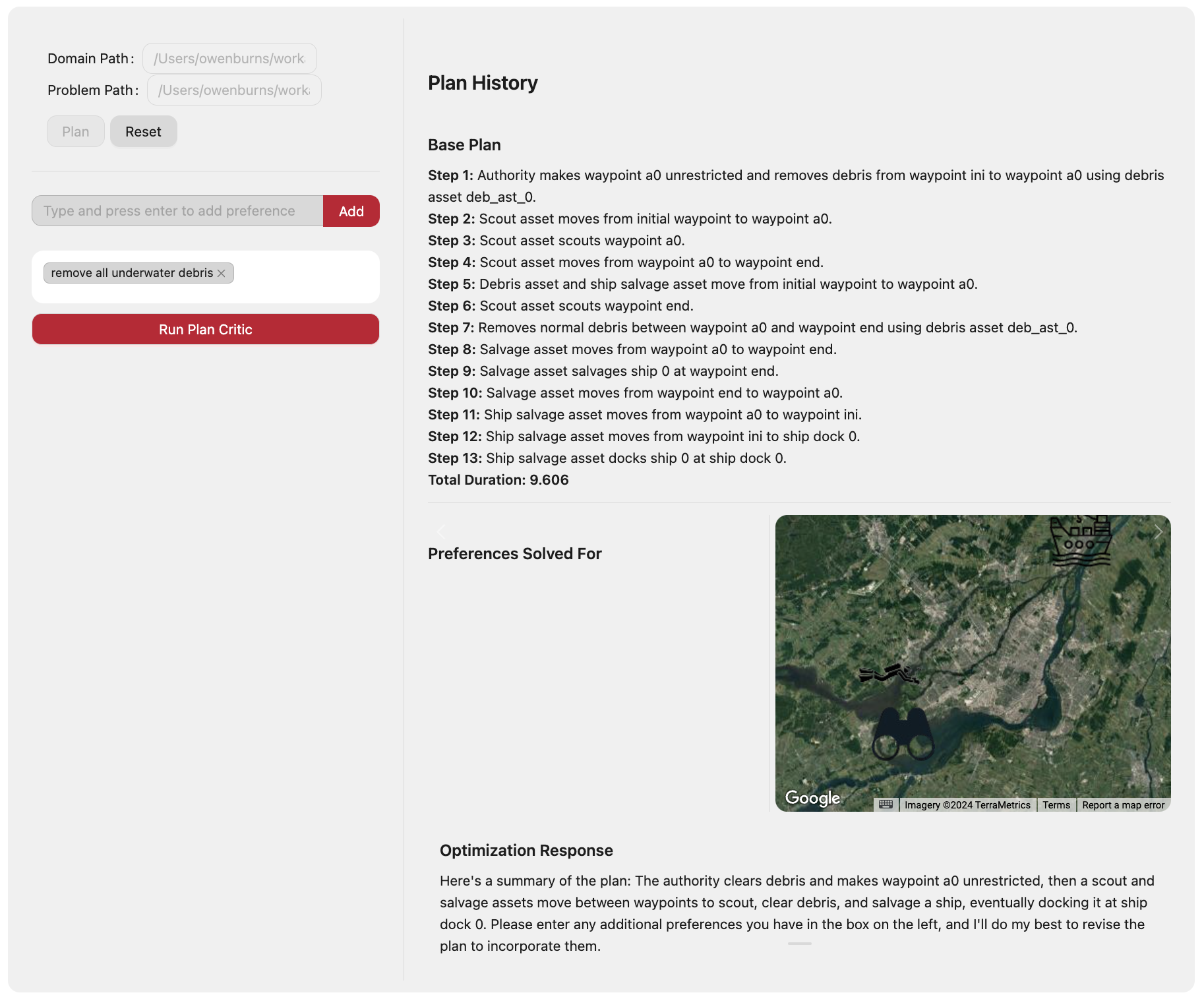}
  \vspace{-.1cm}
  \caption{Example interface demonstrating needed aspects for user-system interaction.}
  \label{fig:user_interface}
\end{figure}

Figure~\ref{fig:user_interface} provides an envisioned interface to provide information to a user by the system.  For a given task, the user is provided an initial plan (as a sequence of steps) generated by the symbolic planner, given an initial problem and goal description, and a summary of the plan.  The user may enter one or more natural language statements as \textit{feedback} about the current plan to indicate additional preferences he or she would prefer about the plan.  We denote the feedback as a set of individual statements, $F = \{f_1, f_2, ..., f_n\}$.

The system utilizes an LLM to initially translate each feedback statement, $f_i$, into an atomic PDDL constraint, $c_i$ (defined in Section~\ref{sec:planner}).  The LLM first translates the feedback statement into a \textit{mid-level constraint}, which is a a natural language constraint that is grounded in the symbolic objects in the PDDL problem description.  Table~\ref{table:symbolic_grounding} provides examples of \textit{mid-level constraints} for provided individual feedback used in the scenario described in~\ref{sec:evaluation}.  We used GPT-4 as the LLM to perform initial translation, as well as translating individual plan steps to natural language, and generating a summary of the plan from plan steps.

\begin{table*}[!tb]
\caption{Mid-level constraints generated from example feedback for the naval disaster response scenario.}
\begin{adjustbox}{width=\textwidth}
\begin{tabular}{|p{0.3\linewidth}|p{0.7\linewidth}|}
\hline
\textbf{Individual Feedback} & \textbf{Mid-Level Constraint} \\
\hline
\makecell[l]{Make sure the scout asset \\ only visits the endpoint once} & \makecell[l]{Limit the scout asset (`sct\_ast\_0`) to visiting the endpoint (`wpt\_end`) at most \\ one time throughout the entire plan.} \\
\hline
\makecell[l]{We need to clear the route \\ from debris station 0 to the \\ endpoint within 5 hours} & \makecell[l]{Ensure that after time step 5, the route between `deb\_stn\_0` and `wpt\_end` is always \\ unblocked.} \\
\hline
\makecell[l]{Don't remove any \\ underwater debris} & \makecell[l]{Ensure that the underwater debris u\_deb\_ini\_b\_0 remains at wpt\_ini at all times. \\ Ensure that the underwater debris u\_deb\_b\_0\_end remains at wpt\_b\_0 at all times.} \\
\hline
\end{tabular}
\end{adjustbox}
\label{table:symbolic_grounding}
\end{table*}

\subsection{Symbolic Planner}\label{sec:planner}
%\dth{\textbf{[Owen:  Is the system given an initial goal description, or do specifications entail the "main" goal?]}
Our system is provided with the \textit{domain description}, \textit{problem description}, and \textit{goal description} of a task to generate a plan for in PDDL.  We define a \textit{plan specification}, $S$ as a conjunction of atomic PDDL constraints

\begin{equation}
    S = \bigwedge\limits_{i=1}^{|S|} c_i
\end{equation}

\noindent where $c_i$ denotes an atomic PDDL \textit{state-trajectory} constraint that corresponds to an individual feedback statement, $f_i$.  PDDL constraints are defined by the grammar in Figure~\ref{fig:grammar}.  $<\!pred\!>$ indicates a valid predicate in the domain and problem description, $<\!dur\!>$ indicates a valid time (or step number) during the plan trajectory, and $<\!cond\!>$ indicates a logical statement consisting of one or more predicates.  $c_i$ encapsulates all state-trajectory constraints and corresponding semantics available in PDDL3 that we used.  $|S|$ indicates the number of atomic constraints in the specification.

\begin{figure}[htbp]
\begin{equation*}
\begin{aligned}
    % c_i~::= &~always~<\!pred\!> \\
    %        | &~always~(not~<\!pred\!>) \\
    %        | &~always~(or~<\!pred\!>~<\!pred\!>) \\
    %        | &~always~(and~<\!pred\!>~<\!pred\!>) \\
    %        | &~sometime~<\!pred\!> \\
    %        | &~sometime~(or~<\!pred\!>~<\!pred\!>) \\
    %        | &~at\text{-}most\text{-}once~<\!pred\!> \\
    %        | &~at\text{-}most\text{-}once~(or~<\!pred\!>~<\!pred\!>) \\
    %        | &~sometime\text{-}before~<\!pred\!>~<\!pred\!> \\
    %        | &~sometime\text{-}after~<\!pred\!>~<\!pred\!> \\
    %        | &~at~end~<\!pred\!> \\
    %        | &~at~end~(not~<\!pred\!>) \\
    %        | &~at~end~(or~<\!pred\!>~<\!pred\!>) \\
    %        | &~always\text{-}within~<\!dur\!>~<\!pred\!> \\
    %        | &~hold\text{-}during~<\!dur\!>~<\!dur\!>~<\!pred\!> \\
    %        | &~hold\text{-}after~<\!dur\!>~<\!pred\!> \\
    c_i~::= &~always~<\!cond\!> \\
           | &~sometime~<\!cond\!> \\
           | &~within~<\!dur\!>~<\!cond\!> \\
           | &~at\text{-}most\text{-}once~<\!cond\!> \\
           | &~sometime\text{-}after~<\!cond\!>~<\!cond\!> \\
           | &~sometime\text{-}before~<\!cond\!>~<\!cond\!> \\
           | &~always\text{-}within~<\!dur\!>~<\!cond\!> \\
           | &~hold\text{-}during~<\!dur\!>~<\!dur\!>~<\!cond\!> \\
           | &~hold\text{-}after~<\!dur\!>~<\!cond\!> \\
           | &~at~end~<\!cond\!> \\
    <\!cond\!>~::= &(<\!pred\!>) \\
           | &(not~<\!pred\!>) \\
           | &(or~<\!cond\!>~<\!cond\!>) \\
           | &(and~<\!cond\!>~<\!cond\!>) \\
\end{aligned}
\end{equation*}
\caption{Grammar used to define feedback as PDDL constraints.}
\label{fig:grammar}
\end{figure}

Our system utilizes an arbitrary symbolic planner to generate a plan, $P$, using the provided domain and problem description, and conjoining the plan specification $S$ to the initial goal descriptions.  A plan consists of a sequence of actions, $P = (a_0, a_1, ..., a_n)$; plan length is denoted as $|P|$.   For our experiments, we utilized the OPTIC (Optimizing Preferences and TIme-dependent Costs) planner~\cite{benton2012temporal}.

\subsection{Specification Adherence Model}\label{sec:adherence_model}
Our system utilizes a \textit{Specification Adherence Model} as a means to quantify the extent to which a generated plan adheres to a set of user feedback statements.  This model takes a plan, $P$ and user feedback statement, $f_i$ as input, and generates the following output, $\nu$ indicating if the plan adheres to or violates the given feedback statement

\begin{equation}
\nu(P,f_i) =
\begin{cases}
    1, \quad \text{if $P$ adheres to $f_i$} \\
    0, \quad \text{if $P$ violates $f_i$}
\end{cases}
\end{equation}

From this, the \textit{adherence rate} of a plan to user feedback, $R(P,F)$ can be calculated as

\begin{equation}
    R(P,F) = \frac{\sum_{i=1}^{|S|} \nu(P,f_i) }{|S|}
\end{equation}

We represented $\nu(P,f_i)$ as a neural network consisting of a Long Short-Term Memory (LSTM) layer of 512 units, followed by two fully connected layers of 256 units with rectified linear (ReLU) activations; the output layer consisted of a single output with a sigmoid activation.  Dropout between the two fully connected layers was utilized during training with a dropout rate of 0.5.

To evaluate the adherence of the pal to a feedback statement, each step of the plan and the feedback statement are embedded using OpenAI's \texttt{text-embedding-3-small} model\footnote{\url{https://openai.com/index/new-embedding-models-and-api-updates/}}.  The embedding of each plan step and the embedding of the feedback statement are concatenated, and the sequence of plan step / feedback embeddings is provided as input to the network.  The plan is considered to adhere to the feedback if the output of the model exceeds 0.5.

\subsection{Genetic Algorithm Component}\label{sec:genetic_algorithm}
The initial plan specifications generated by the LLM from user feedback statements are prone to mistranslation.  In order to generate a plan with improved adherence to the user's feedback, the system uses a genetic algorithm to search for plan specifications which produce adherent plans.

The genetic algorithm maintains a population of size $M$; each individual $i$ in the population is defined by a plan specification, $S_i$.  We denote the population as $\mathcal{P}^{k} = \{ S^{k}_1, S^{k}_2, ... S^{k}_M \}$, where $k$ denotes the generation number.

The genetic algorithm produces a new population from a current population using mutation and crossover operations.  The mutation operation, summarized in Algorithm~\ref{alg:mutation}, involves modifying a given plan specification by i) adding a random atomic constraint, ii) removing an existing atomic constraint, iii) randomly negating or changing the arguments of the constraint's predicates or changing the state trajectory constraint used, or iv) duplicating an existing atomic constraint and applying one of the aforementioned random modifications.  The crossover operation, summarized in Algorithm~\ref{alg:crossover}, involves exchanging atomic constraints in two parent specifications at a random crossover point.

\begin{algorithm}[htb]
    \caption{Mutation Operation}
    \label{alg:mutation}
    \textbf{Input}: $S^{k} := \{c_1, c_2, \ldots c_n\}$ \\
    \textbf{Output}: $S^{k+1}$
    \begin{algorithmic}[1] %[1] enables line numbers
        \STATE $type \sim \{ add, remove, modify \}$.
        \IF{$type = add$}
            \STATE Let $c_{n+1} \gets $ random constraint
            \STATE $S^{k+1} \gets S^{k} \cup c_{n+1}$
        \ELSIF{$type = remove$ and $|S^{k} > 1|$}
            \STATE Let $j \sim \{1 \ldots n\}$
            \STATE $S^{k+1} \gets S^{k} - \{c_j\}$
        \ELSE
            \STATE Let $j \sim \{1 \ldots n\}$
            \STATE $rule \sim \{negate, state\text{-}trajectory, predicate\}$
            \IF{$rule = negate$}
                \STATE $c'_j \gets not~c_j$
            \ELSIF{$rule = state\text{-}trajectory$}
                \STATE $c'_j \gets c_j$
                \STATE $c'_j[st\text{-}constraint] \sim \{always, sometimes, \ldots\}$
            \ELSE
                \STATE $c'_j \gets c_j$
                \STATE $c'_j[predicate] \gets ChangeArgument(predicate)$
            \ENDIF
        \STATE $S^{k+1} \gets S^{k} \cup {c'_j} - {c_j} $
        \ENDIF
        \RETURN $S^{k+1}$
    \end{algorithmic}
\end{algorithm}

\begin{algorithm}[htb]
    \caption{Crossover Operation}
    \label{alg:crossover}
    \textbf{Input}: $S^{k}_i := \{c^i_1, c^i_2, \ldots c^i_n\}, S^{k}_j := \{c^j_1, c^j_2, \ldots c^j_m\}$ \\
    \textbf{Output}: $S^{k+1}_i, S^{k+1}_j$
    \begin{algorithmic}[1] %[1] enables line numbers
        \STATE $p \sim \{1 \ldots min(m,n)\}$
        \STATE $S^{k+1}_i \gets S^{k}_i[1:p] \cup S^{k}_j[p:m]$
        \STATE $S^{k+1}_j \gets S^{k}_j[1:p] \cup S^{k}_i[p:n]$        
        \RETURN $S^{k+1}_i, S^{k+1}_j$
    \end{algorithmic}
\end{algorithm}

To evaluate the fitness of an individual in the population, $o^{(k)}_i$ the symbolic planner is used to generate the plan, $P^{(k)}_i$, with corresponds to the individual's plan specification, $S^{(k)}_i$.  The fitness of the individual is calculated as the adherence rate of the generated plan to the user feedback,

\begin{equation}
    o^{(k)}_i = R(P^{(k)}_i, F)
\end{equation}

At each generation, 50\% of the population with the highest fitness score are maintained as an elite set for the next generation.  Parents are selected randomly to generate children for the next generation, and the generated children provide the remaining 50\% of the population for the next generation.  We used a population size of 20, and run the genetic algorithm for three generations or until an adherence score of 100\% is achieved.   The genotypes of the initial population were generated by performing a mutation operation on the initial atomic constraint generated by the LLM.

%%%%%%%%%%%%%%%%%%%%%%%%%%%%%%%%%%%%%%%%%%%%

\section{EVALUATION}\label{sec:evaluation}
We developed a simulated naval disaster response scenario, and constructed a set of natural language \textit{constraint archetypes} as a basis for natural language feedback statements. 

\subsection{Scenario Description}
\begin{figure}[!h]
  \centering  \includegraphics[width=0.7\linewidth]{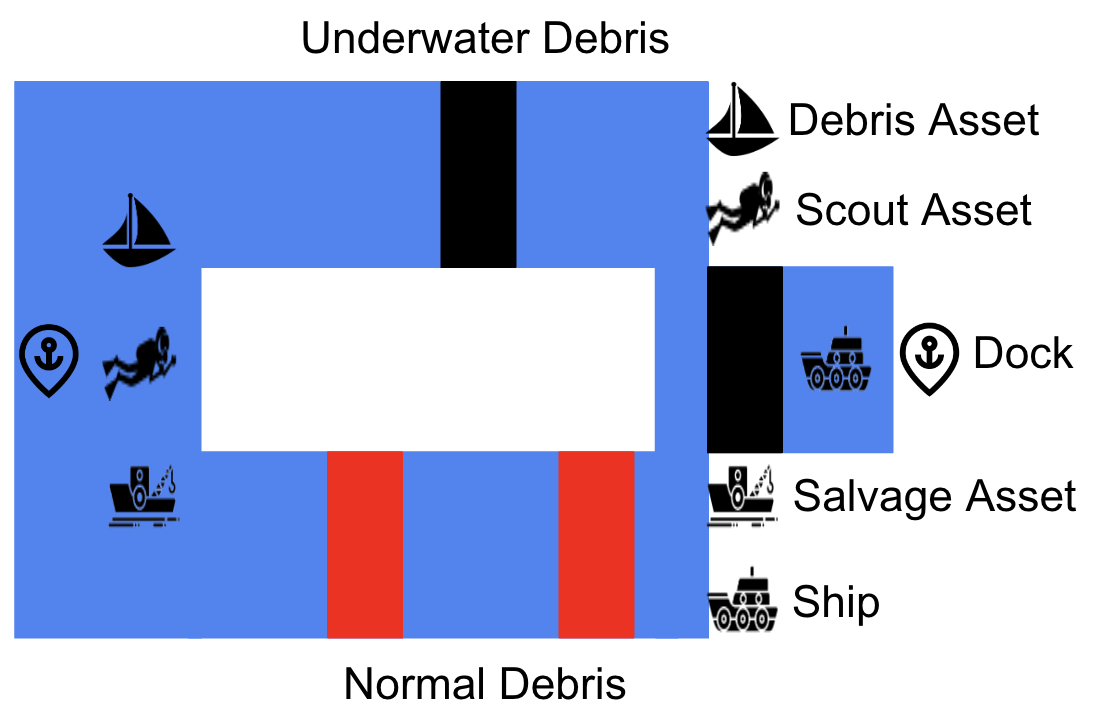}
%  \vspace{-.5cm}
  \caption{Naval disaster response scenario, showing two waterways separating assets from a target ship to salvage.  Waterways are blocked by underwater debris (black), or floating debris (red), that must be removed prior to traversing the waterway. Underwater debris must be discovered by the scout asset before removal.}
  \label{fig:environment}
\end{figure}

The naval disaster response scenario reflects a post-disaster scenario in which both floating and underwater debris may be blocking access through waterways.  In our scenario, a notional decision maker is tasked with removing debris from the waterway in order to move a derelict ship from a dock to a target location.  Figure~\ref{fig:environment} provides an illustration of the scenario used for evaluation.  In the scenario, the following symbolic objects types are defined:

\begin{itemize}
    \item \textbf{Waypoint.}  Defining locations in the environment.
    \item \textbf{Normal Debris.}  Debris removable by a Debris Asset.
    \item \textbf{Underwater Debris.}  Debris that must be discovered by a Scout Asset prior to removal by a Debris Asset.
    \item \textbf{Debris Asset.}  A vessel capable of collecting debris at a waypoint, or depositing debris at (another) waypoint.
    \item \textbf{Scout Asset.}  An asset that discovers underwater debris.
    \item \textbf{Ship.}  Derelict ship initially located at a \textit{Dock} waypoint.
    \item \textbf{Salvage Asset.}  An asset that is capable of moving the Ship from one waypoint to another.
\end{itemize}

In the scenario, the \textit{Ship} is initially located at a defined \textit{Dock} waypoint, and the initial goal description indicates that the \textit{Ship} is located at a target waypoint.  In order to succeed at the goal, a plan must clear debris from a waterway, so that the \textit{Salvage Asset} can move the \textit{Ship} from the \textit{Dock} to the target waypoint.   Assets can travel between connected waypoints, but cannot travel from a waypoint which contains debris until the debris is removed.

\subsection{Specification Adherence Training}

To construct a dataset to train the specification adherence model, we first created four variations of the waterway restoration \textit{problem description} and programmatically generated the full set of possible atomic \textit{state-trajectory constraints} afforded by the \textit{domain description}. For each problem variation, we generate ~1000 \textit{feedback instances}, each consisting of a plan and a set of natural language preferences split evenly between positive and negative examples.

To generate a feedback instance, we create problem specifications consisting of 2-5 constraints sampled from the set of possible atomic constraints and test them for solvability using Optic. If a plan is generated successfully, we then sample a second problem specification with an equal number of constraints, and use VAL \cite{1374201} to verify that the previously generated plan \textbf{does not} adhere to any of them. These problem specifications form the positive and negative set, respectively, and are translated from PDDL to natural language using an LLM to match the semantics which would be encountered in practice.
Our final dataset consisted of 3,422 feedback instances containing a total of 22,250 examples.
We trained the model using binary cross entropy loss \cite{mao_cross-entropy_2023} for 100 epochs, achieving a final validation loss of 0.49 and validation accuracy of 75\%.

\subsection{Evaluation Dataset Generation}

For the described scenario, we constructed a set of \textit{constraint archetypes} in natural language, and the corresponding ground-truth \textit{plan specifications}.  For each constraint archetype, we used an LLM to create ~30 rephrasings of each archetype for use as an individual feedback element. Our final dataset consisted of 277 feedback elements.

\section{RESULTS}\label{sec:results}

To evaluate the effectiveness of our approach, we generated a plan for each natural language feedback element in the evaluation dataset, and used VAL to determine if the generated plan adhered to the ground-truth \textit{plan specification} corresponding to the plan archetype the feedback element was derived from. To determine the influence of the genetic algorithm and specification adherence model, we also determined the number of valid plans generated using \textit{only} the initial LLM translation of \textit{feedback} to \textit{plan specification}.

Table \ref{table:goals} shows the results of our system on the generated evaluation dataset.  Our system was able to generate valid plans for given feedback at a rate of 47.65\%; for comparison, the LLM-only system achieved at a rate of 32.49\%.

\begin{figure}[!ht]
  \centering  \includegraphics[width=0.7\linewidth]{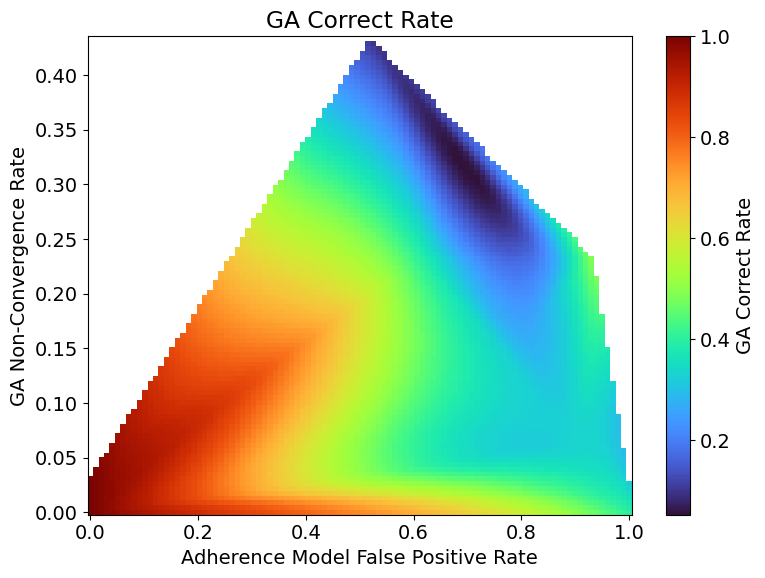}
%  \vspace{-.5cm}
%  \caption{GA success diminishes plotted against GA convergence failure rate and adherence model false positive rate. }
  \caption{Overall GA success rate based on the non-convergence rate of the GA (y-axis) and failures due to false positives of the adherence model (x-axis).}
  % for cases where 3 generations of mutation are insufficient to reach a valid specification, where the adherence model returns false positives, or both.

  \label{improvement_methods}
\end{figure}

\begin{table}[h!]
\centering
\caption{Number of valid and invalid plans generated by our system and LLM-only translation for each constraint archetype.  Bolded entries indicate higher rate of valid plans.}
\begin{tabular}{|p{3cm}|cc|cc|}
\hline
\textbf{Constraint} & \multicolumn{2}{c|}{\textbf{Our Approach}} & \multicolumn{2}{c|}{\textbf{LLM-Only}} \\ \cline{2-5} 
\textbf{Archetype} & \textbf{valid} & \textbf{invalid} & \textbf{valid} & \textbf{invalid} \\ \hline
All underwater debris is removed & 12 & 18 & 12 & 18 \\ \hline
Scout asset reaches end point before debris asset moves & \textbf{30} & 0 & 16 & 14 \\ \hline
Waypoint b is made unrestricted & \textbf{12} & 18 & 3 & 27 \\ \hline
Step 6 happens before step 5 & \textbf{10} & 20 & 0 & 30 \\ \hline
All of the underwater debris is removed and none of the normal debris is removed & \textbf{12} & 25 & 3 & 34 \\ \hline
Debris asset ends at waypoint b & 8 & 22 & \textbf{9} & 21 \\ \hline
All assets are at the ship dock at the end of the plan & 3 & 27 & \textbf{28} & 2 \\ \hline
Scout asset reaches shipwreck before debris asset reaches shipwreck & \textbf{30} & 0 & 14 & 16 \\ \hline
Scout asset reaches shipwreck before debris asset reaches shipwreck and no underwater debris is removed & \textbf{15} & 15 & 5 & 25 \\ 
\hline
\end{tabular}
\label{table:goals}
\end{table}

Table \ref{table:correctness_naval} compares the results of our system and the LLM-only system.  We find that the genetic algorithm is adept at correcting the LLM when it gives an incorrect translation, returning a correct plan 40\% of the time when the feedback translation generated by the LLM is either incorrect or results in a failure to generate a plan.

Figure~\ref{improvement_methods} illustrates two principal failure modes that degrade the performance of our approach: (1) failure of the GA to converge on a valid plan specification within three generations (y-axis), and (2) false-positive adherence judgments by the LSTM-based adherence model (x-axis).  The first failure mode arises with constraint archetypes that mandate extended plan sequences. For instance, enforcing that “all assets end at the ship dock” requires at least three additional actions to relocate ships to the dock, resulting in a 43\% non-convergence rate for this archetype. The second failure mode is evident in constraints that include multiple semantically similar requirements, e.g., ``all underwater debris is removed'' can appear once or multiple times. Distinguishing these variations proved challenging for the adherence model, causing a 100\% false-positive rate for this archetype.

Additional analysis of the failure cases indicate that the average length of generated plans are roughly equivalent for successful cases, cases where the GA failed to converge, and failures due to false positives from the adherence model.  Additionally, the rates of each type of mutation performed in successful and failure cases were roughly equivalent, with the exception that cases with false positives of the adherence models exhibited a higher rate of changing constraint types ($22.5\%$) compared to success cases ($12.7\%)$.  In particular, most failures occur with constraints $always$ or $at~end$ ($53.8\%$ and $66.1\%$ of non-convergent and adherence model false positives, respectively).  We note that these types of constraints correspond to the archetypes that the model failed to improve upon (e.g., "Debris asset ends at waypoint b", "All assets are at the ship dock at the end of the plan").

\begin{table}[h!]
\centering
\caption{Cross-comparison of LLM and GA performance\\%
         \makebox[\linewidth][c]{Naval disaster-response scenario}}
\begin{tabular}{|l|c|c|c|}
\hline
 \multicolumn{2}{|c|}{ } & \multicolumn{2}{c|}{Our Approach} \\
\cline{3-4}
 \multicolumn{2}{|c|}{ } & \textbf{valid} & \textbf{invalid} \\ \hline
\multirow{2}{*}{LLM} & \textbf{valid} & 56 & 34 \\
\cline{2-4}
& \textbf{invalid} & 76 & 111 \\ \hline
\end{tabular}
\label{table:correctness_naval}
\end{table}

\begin{table}[h!]
\centering
\caption{Cross-comparison of LLM and GA performance\\%
         \makebox[\linewidth][c]{Satellite scenario}}
\begin{tabular}{|l|c|c|c|}
\hline
 \multicolumn{2}{|c|}{ } & \multicolumn{2}{c|}{Our Approach} \\
\cline{3-4}
 \multicolumn{2}{|c|}{ } & \textbf{valid} & \textbf{invalid} \\ \hline
\multirow{2}{*}{LLM} & \textbf{valid} & 7 & 2 \\
\cline{2-4}
& \textbf{invalid} & 40 & 91 \\ \hline
\end{tabular}
\label{table:correctness_satellite}
\end{table}

\subsection{Validation domain}
To reinforce our results, we ran the same set of experiments on the simple time version of the Satellite domain from 2002 ICP planning competition \cite{Long_2003}. Problems derived from this domain represent the task of planning sensor usage to take a set of measurements in the most efficient order. Various types of sensors are involved requiring the plan to balance both the order of calibrating the sensors and the movement of the satellite to vantage points from which the measurements can be taken.

Table \ref{table:correctness_satellite} shows the result of this experiment. In this case, our system generated valid plans for feedback at a rate of 33.57\%, while the LLM-only system achieved a rate of 6.43\%. While the overall accuracy of the system is lower than on the naval planning domain, the improvement over the LLM-only baseline is clear. 

\section{CONCLUSION AND FUTURE WORK}\label{sec:conclusion}
We propose a neurosymbolic architecture to optimize PDDL plans with respect to user preferences in online and dynamic scenarios. We use a genetic algorithm to evolve the initial translation provided by an LLM into a more adherent problem specification. We find that our approach is superior at generating plans whose state trajectory aligns with stated user preferences than a neurosymbolic architecture using an LLM alone, and that it is extremely effective at catching the LLMs mistakes. 
However, performance degrades in cases that require extended plan horizons or involve multiple semantically similar but disjoint actions.  
%\dth{Additionally, performance degredation appears to occur in cases that involve $always$ and $at~end$ constraint types; improving the adherence model's ability to correctly categorize plan ahderence to these constraints should improve overall performance of the approach.}

Future work includes testing out different architectures of reward model, as well as measuring how overall performance changes when using a smaller LLM (e.g. GPT-3.5) to generate the initial candidate. Potential avenues for improving the performance of the GA include augmenting the adherence model training dataset and increasing the number of GA iterations. We intend to expand the training dataset to include natural language feedback derived from multiple constraints from the problem specifications. This will ensure that cases where the feedback is partially satisfied are included, allowing the specification adherence model to learn to effectively handle the cases which caused false positives in our experiment.

\addtolength{\textheight}{-0cm}   % This command serves to balance the column lengths

%\appendix

%\section*{Ethical Statement}

%There are no ethical issues.

%\section*{ACKNOWLEDGEMENT}
%This project was funded by ONR Award No. N000142312840 and NSF Award No. 2021-67021-35329.

\bibliographystyle{ieeetr}
\bibliography{citations}

\end{document}